%% file: main.tex
\documentclass[10pt,twocolumn,letterpaper]{article}

\usepackage{iccv}              
\usepackage{amsmath}
\usepackage{algorithmic}
\usepackage{algorithm}
\usepackage{textcomp}
\usepackage{graphicx}
\usepackage{multirow}
\usepackage{booktabs}
\usepackage{caption}
\usepackage{subcaption}
\usepackage{tikz}
\usetikzlibrary{bayesnet}


\definecolor{iccvblue}{rgb}{0.21,0.49,0.74}
\usepackage[pagebackref,breaklinks,colorlinks,allcolors=iccvblue]{hyperref}

\def\paperID{*****} 
\def\confName{ICCV}
\def\confYear{2025}

\title{Dataset Distillation with Probabilistic Latent Features}

\author{Zhe Li\\
FAU Erlangen-Nürnberg \\
{\tt\small zhe.li@fau.de}
\and
Sarah Cechnicka\\
Imperial College London\\
{\tt\small sarah.cechnicka18@imperial.ac.uk}
\and
Cheng Ouyang\\
Imperial College London\\
{\tt\small c.ouyang@imperial.ac.uk}
\and
Katharina Breininger\\
FAU Erlangen-Nürnberg \\
{\tt\small katharina.breininger@fau.de}
\and
Peter Sch\"uffler\\
Technical University Munich\\
{\tt\small peter.schueffler@tum.de}
\and
Bernhard Kainz\\
FAU Erlangen-Nürnberg \\
{\tt\small bernhard.kainz@fau.de}
}

\begin{document}
\maketitle
\input{sec/0_abstract}    
\input{sec/1_intro}

\input{sec/2_method}

\input{sec/3_experiment}

{
    \small
    \bibliographystyle{ieeenat_fullname}
    \bibliography{main}
}

\input{sec/X_suppl}

\end{document}

%% file: sec/0_abstract.tex
\begin{abstract}
As deep learning models grow in complexity and the volume of training data increases, reducing storage and computational costs becomes increasingly important. Dataset distillation addresses this challenge by synthesizing a compact set of synthetic data that can effectively replace the original dataset in downstream classification tasks. While existing methods typically rely on mapping data from pixel space to the latent space of a generative model, we propose a novel stochastic approach that models the joint distribution of latent features. This allows our method to better capture spatial structures and produce diverse synthetic samples, which benefits model training.
Specifically, we introduce a low-rank multivariate normal distribution parameterized by a lightweight network. This design maintains low computational complexity and is compatible with various matching networks used in dataset distillation. After distillation, synthetic images are generated by feeding the learned latent features into a pretrained generator. These synthetic images are then used to train classification models, and performance is evaluated on real test set.
We validate our method on several benchmarks, including ImageNet subsets, CIFAR-10, and the MedMNIST histopathological dataset. Our approach achieves state-of-the-art cross architecture performance across a range of backbone architectures, demonstrating its generality and effectiveness.
\end{abstract}

%% file: sec/1_intro.tex
\section{Introduction}
\label{sec:intro}
As data volumes continue to grow exponentially, training increasingly over-parameterized deep learning models incurs significant storage and computational costs. Dataset distillation addresses this challenge by synthesizing a compact set of representative images that encapsulates the essential characteristics of a larger dataset, enabling downstream tasks to achieve performance comparable to models trained on the full original dataset.

Early methods in dataset distillation primarily operated directly in the pixel space~\cite{zhao2020dataset,zhao2021dataset,cazenavette2022dataset,zhao2023DM,zhang2023accelerating,zhao2023improved}, but these approaches often introduced noise and artifacts, especially in high-resolution images. Recent strategies, such as GLaD~\cite{cazenavette2023generalizing}, have shifted attention to latent space distillation, leveraging pretrained generative models such as StyleGAN-XL~\cite{sauer2022stylegan} as deep generative priors. While latent space methods substantially reduce noise compared to pixel-space methods, they still struggle with residual artifacts due to insufficient consideration of spatial correlations and structural coherence within images.

In natural images, spatial correlation and structural coherence are fundamental, as neighboring pixels generally exhibit similar attributes and consistent semantic information. Existing distillation methods typically update images or latent representations through standard backpropagation without explicitly modeling these spatial relationships. Such an approach neglects the inherent ambiguity of the dataset distillation process, where multiple valid synthetic representations might exist. Effectively modeling this inherent uncertainty could significantly improve the distillation outcomes by guiding synthetic image generation toward structurally consistent and semantically meaningful results.
Uncertainty in image data typically comprises aleatoric uncertainty, which stems from inherent variability in the observations, and epistemic uncertainty, arising from incomplete knowledge or limited data~\cite{kendall2017uncertainties}. Specifically, aleatoric uncertainty in images is often spatially correlated and exhibits heteroscedasticity, reflecting varied uncertainty across different regions within the same image~\cite{monteiro2020stochastic}. However, pretrained generative models like StyleGAN-XL typically produce deterministic and spatially independent outputs, limiting their ability to fully capture structured uncertainty.

To address this limitation, we introduce the Stochastic Latent Feature Distillation (SLFD) framework. SLFD explicitly models spatial correlations and uncertainty within latent features by employing a low-rank multivariate normal distribution. Given that synthetic images are ultimately produced from latent features passed through a pretrained generator, we assume that directly modeling uncertainty within the latent feature space provides an efficient and effective approach. The stochastic component of our method generates multiple latent feature samples, explicitly embedding spatial coherence and uncertainty into the distillation process. Additionally, our proposed module is compact and comprises only three linear layers, ensuring compatibility with existing matching algorithms without necessitating architectural modifications. 

Beyond natural image datasets, dataset distillation is increasingly valuable in healthcare settings, where data privacy and efficient data sharing are critical~\cite{you2023bootstrapping,fang2023reliable,kanwal2023vision}. Synthesizing anonymized and compact representations can facilitate secure data sharing among clinical institutions by effectively removing identifiable patient information. In our experiments, we demonstrate the effectiveness of SLFD on histopathological image datasets, highlighting its potential for impactful applications in medical imaging.

Our contributions can be summarized as follows:

\begin{enumerate}
  \item We introduce Stochastic Latent Feature Distillation (SLFD), a novel dataset distillation framework that incorporates structured uncertainty through a low-rank multivariate distribution, enabling the generation of informative and spatially coherent synthetic data.

  \item SLFD effectively models spatial relationships in the latent space and maintains high performance even at large image resolutions, offering robustness and scalability with minimal computational overhead.

  \item Through extensive experiments on CIFAR-10 and multiple ImageNet subsets, SLFD outperforms state-of-the-art methods across diverse architectures, demonstrating strong cross-model generalization.

  \item We validate SLFD on the MedMNIST histopathology dataset, where it shows consistent quantitative improvements and clear qualitative distinctions, proving its adaptability to complex real-world domains.
  
\end{enumerate}

\section{Related work}

\subsection{Dataset distillation} 
A comprehensive overview over dataset distillation is provided in recent surveys~\cite{lei2023comprehensive,yu2023dataset}.
After the pioneered~\cite{wang2018dataset}, efficiency has been addressed by Generative Teaching Networks (GTNs)~\cite{such2020generative}
and through model regularization techniques~\cite{nguyen2020dataset}. 
The field diversified into various matching methods to align the original and the synthesized data, such as Dataset Condensation with Gradient Matching (DC)~\cite{zhao2020dataset,zhao2021dataset,zhang2023accelerating}, Distribution Matching (DM)~\cite{zhao2023DM,zhao2023improved}, and Matching Training Trajectories (MTT)~\cite{cazenavette2022dataset}.
Beyond matching strategies, methods to align features of convolutional networks~\cite{wang2022cafe,sajedi2023datadam} have been proposed to improve performance.
\cite{liu2022dataset} enhanced the correlation between generated samples through factorization during training. 
\cite{du2023minimizing} tackled the challenge of accumulated trajectory errors in weight initialization during the evaluation phase by guided flat trajectories during training. 
\cite{zhu2023rethinking} introduced new calibration techniques for deep neural networks to mitigate overconfidence issues and the over-concentration in distillation data.

Recognizing the limitations of pixel space, characterized by high-frequency noise, \cite{zhao2022synthesizing} shifted focus to synthesizing images in the latent space using pre-trained GANs, thereby extracting more informative samples. 
Aiming for simplicity and efficiency, \cite{cazenavette2023generalizing} utilized a pre-trained StyleGAN-XL~\cite{sauer2022stylegan} to create a single synthetic image per class from latent space, streamlining the distillation process from real datasets.
The field continues to evolve with methods addressing various phases of dataset distillation, including the introduction of the distillation space concept~\cite{liu2023fewshot} and the implementation of a clustering process for selecting mini-batch real images~\cite{liu2023dream}.

\subsection{Stochastic mapping} 
Bayesian methods have experienced significant attention for neural network robustness, uncertainty estimation, and model regularization.
In classification, significant efforts focus on predicting Dirichlet distributions~\cite{malinin2018predictive,malinin2019reverse,sensoy2018evidential} and post-training calibration of predicted class probabilities~\cite{guo2017calibration}. 
Other approaches~\cite{wilson2020bayesian} and effective approximations~\cite{monteiro2020stochastic} have highlighted the importance of well-calibrated uncertainty estimates in deep neural networks, especially in applications like medical diagnosis, where decision-making under uncertainty is crucial. 

\subsection{Medical applications} 
For medical images, dataset distillation is attractive from a privacy and data sharing perspective. For histopathology images, ~\cite{li2024image} works on MedMnist dataset by a community detection method.
For other medical datasets, the fundamental dataset distillation framework has been applied on gastric X-ray images~\cite{li2020soft,li2022compressed} and a COVID-19 chest X-ray dataset~\cite{li2022dataset}.

%% file: sec/2_method.tex
\begin{figure*}[tb]
   \centering
      \includegraphics[width=1\linewidth]{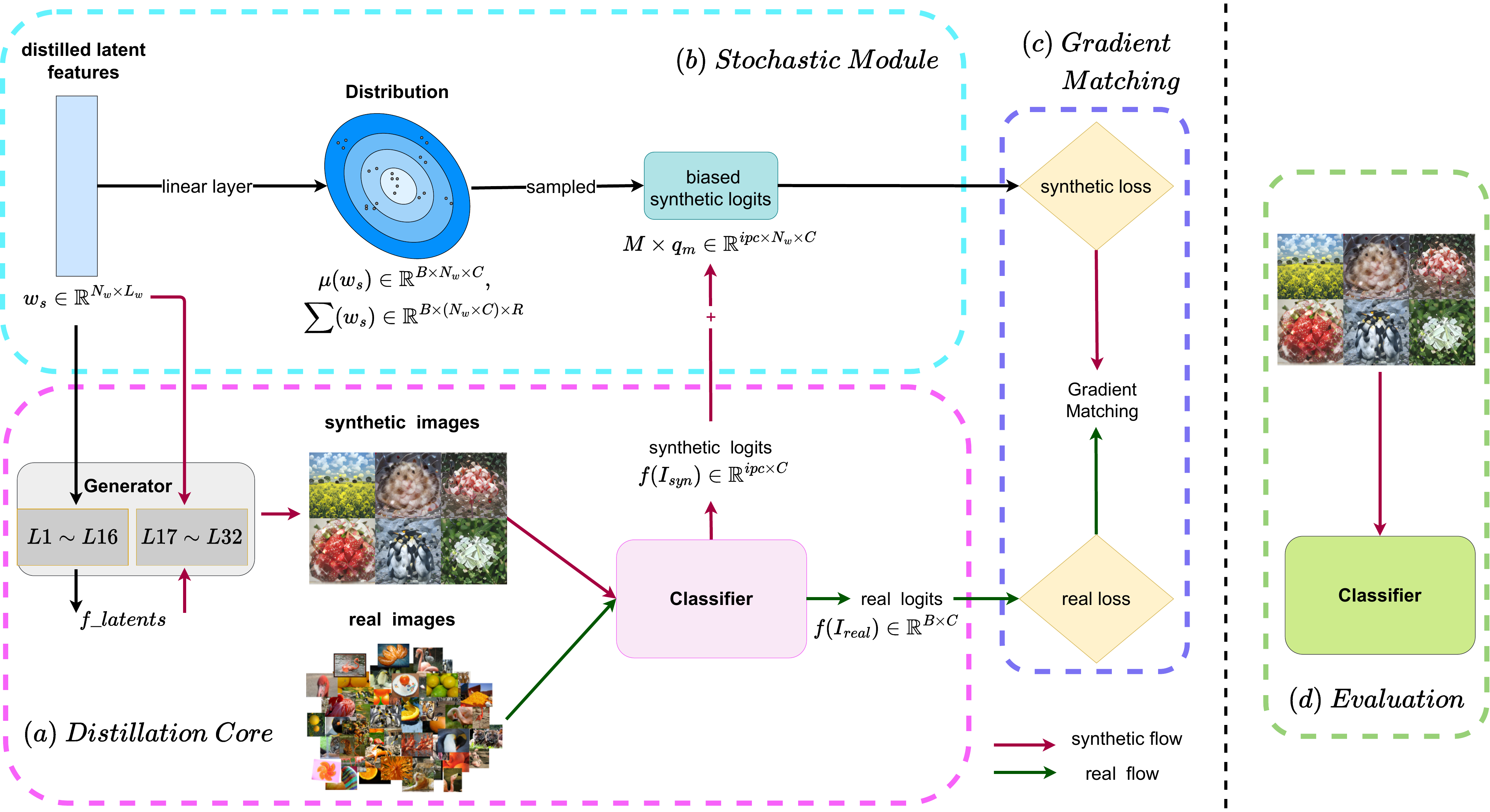}
   \caption{Overview of the proposed SLFD framework. (a) The distillation backbone is based on GLaD~\cite{cazenavette2023generalizing}, where the generator (shown in gray to indicate frozen weights) is split into two parts: the first 16 layers compute intermediate features $f_n$ from latent codes $w_s$ (black arrows), and the remaining 16 layers generate synthetic images using both $w_s$ and $f_n$ (red arrows). (b) The Stochastic Module models uncertainty in the latent space by estimating a low-rank multivariate normal distribution using three linear layers to compute the mean, variance, and covariance factors. Multiple samples drawn from this distribution are used to represent synthetic probability distributions. (c) The Gradient Matching component computes the matching loss between gradients of real and synthetic data. (d) In the Evaluation phase, classifiers are trained solely on the distilled synthetic images.
   }
   \label{fig:overviewframework}
\end{figure*}

\section{Method}
This section introduces the proposed Stochastic Latent Feature Distillation (SLFD) framework, with an overview provided in Fig.\ref{fig:overviewframework}. 
The core of SLFD is a lightweight stochastic module (Fig.\ref{fig:overviewframework}($b$)), described in detail in Section\ref{sec:stochastic}. This module learns a structured distribution over latent features, enabling the generation of diverse and spatially consistent synthetic samples during the distillation process.
Section~\ref{sec:lossfunction} and Fig.~\ref{fig:overviewframework}($c$) explain how we define a stochastic loss function that incorporates samples from the learned latent distribution. This loss guides the optimization process to produce synthetic data that more faithfully preserves the structure and variability of the original dataset.
Finally, as shown in Fig.~\ref{fig:overviewframework}($d$), we assess the quality of the distilled dataset by training independent classifiers solely on the synthetic images.

\subsection{Preliminaries}
\label{sec:preliminary}

\paragraph{Problem introduction} Dataset distillation is a process where the  information contained within a large real dataset $\mathcal{T}$ is ideally condensed into a significantly smaller synthetic dataset $\mathcal{S}$. 
Formally, given a real dataset $\mathcal{T}=\{(x_i, y_i)\}_{i=1}^{N}$, where $x_i\in \mathbb{R}^{3 \times H \times W}$ denotes an image, $y_i \in \{0, 1, 2, ..., C\}$ represents its corresponding class label, in a dataset with $C$ total classes and $N$ total samples. Our task is to synthesize a small synthetic dataset $\mathcal{S}=\{(s_i, y_i)\}_{i=1}^{N_s}$, where $N_s = ipc \times C$, $ipc$ denotes images per class, and $N_s \ll N$. In our experiments, \emph{e.g.}, $ipc=1$ means there is only one image synthesized for each class.

\paragraph{Distillation core}
To mitigate the noise and visual artifacts often introduced by pixel-level distillation methods, GLaD~\cite{cazenavette2023generalizing} performs dataset distillation in the latent space using a pretrained generator as a prior, as illustrated in Fig.~\ref{fig:overviewframework}($a$).
At the beginning of training, latent codes $z_s \in \mathbb{R}^{B \times L_z}$ are sampled from a standard normal distribution $\mathcal{N}(0, 1)$. These codes are then projected into the extended latent space $W^+$~\cite{abdal2019image2stylegan} to produce latent features $w_s \in \mathbb{R}^{B \times N_w \times L_w}$, where $N_w$ denotes the number of style blocks in the StyleGAN-XL mapping network and $L_w$ is the dimensionality of each feature vector.
To enrich the representation and encourage diversity in the synthetic images, the method also extracts intermediate feature activations $f_n$ from a selected hidden layer $n$ within the generator. These features, referred to as $F_n$, provide additional context and structure when synthesizing images.
During the distillation process, synthetic images are generated by passing both $w_s$ and $f_n$ through the latter layers of the frozen StyleGAN-XL generator. The training objective involves matching gradients between synthetic and real images to preserve semantic alignment.
After distillation, the quality of the synthetic dataset is evaluated by training classifiers solely on these generated images, and performance is measured using test accuracy on a real validation set.

\begin{algorithm}[H]
\caption{Our proposed SLFD approach.}\label{algo:psedocodealgo}

\begin{algorithmic}
\STATE \textbf{Input}: The generator $G$, Real dataset $\mathcal{T}$, $C$, $Epoch$, and batch size $B$.
\STATE Initialize latent features $w_s$ and $f_n$ and the Stochastic Module .
\FOR{epoch from 0 to $Epoch$}
    \STATE Synthetic images $\mathcal{S}=G(w_s, f_n)$; Initialize the distillation model $f$.
    \FOR{each class $c$ in $C$ }
        \STATE $ I_{real}^c \sim \text{RandomSubset}(\mathcal{T}_c, $B$)$ $\rightarrow$ $f(I_{real}^c)$ $\rightarrow$ $\mathcal{L}_{real}^c$
        \STATE $I_{syn}^c \subset \mathcal{S}$ $\rightarrow$$f(I_{syn}^c)$ $\rightarrow$ \{$\mu(w_s), \Sigma(w_s)$\} $\rightarrow$ $\{q_m\}_{m=1}^{M}$
        \STATE Update $q_m^* = q_m - \mu(w_s) +  f(I_{syn}^c) $
        \STATE Calculate $\mathcal{L}_{syn}^c$ 
        \STATE Gradient matching $\mathcal{L} = 1 - \frac{\nabla_\theta \mathcal{L}_{syn}^c\cdot \nabla_\theta \mathcal{L}_{real}^c}{\parallel \nabla_\theta \mathcal{L}_{syn}^c\parallel \parallel \nabla_\theta \mathcal{L}_{real}^c \parallel}$
    \ENDFOR
    \STATE Update the latent features $w_s$ and $f_n$ by backpropagation.
\ENDFOR\ Distillation

\STATE Synthetic images $\mathcal{S}=G(w_s, f_n)$. 
\STATE Train classifiers only on the synthetic images $S$.
\STATE Evaluate all models on large real test set.
\end{algorithmic}
\end{algorithm}

\subsection{Stochastic training}
\label{sec:stochastic}

The motivation for integrating the stochastic module into the distillation framework is to incorporate structured uncertainty into the image generation process, which is typically absent in deterministic models.
Inspired by~\cite{monteiro2020stochastic}, we observe that both the generator and classifier in conventional distillation pipelines behave as deterministic functions. As a result, the synthetic image outputs are treated as independent across pixels, overlooking the spatial dependencies inherent in natural images. To address this, we propose modeling the distribution over the classifier outputs for synthetic images using a multivariate normal distribution:

\begin{equation}\label{multivariateimage}
    f(I_{\text{syn}}) \mid S \sim \mathcal{N}(\mu(S), \Sigma(S))
\end{equation}

Since these synthetic images are generated from latent features, we redefine the distribution in terms of the generator:

\begin{equation}\label{multivariatelatent}
    f(I_{\text{syn}}) \mid G_{16}(w_s; G_0(w_s)) \sim \mathcal{N}(\mu(w_s), \Sigma(w_s))
\end{equation}

where $w_s$ is the latent input, $G_0(w_s)$ produces intermediate features $f_n$ from the first part of the generator, and $G_{16}$ refers to the second half of the generator that produces the image.
Here, $\mu(w_s) \in \mathbb{R}^{B \times N_w \times C}$ is the mean vector, and $\Sigma(w_s) \in \mathbb{R}^{B \times (N_w \times C)^2}$ is the full covariance matrix. The output dimension $C$ corresponds to the number of classes predicted by the classifier, replacing the latent feature dimensionality $L_w$. We use a batch size $B$ equal to $C$, following the convention of one synthetic image per class.

To manage computational complexity, we approximate the full covariance with a low-rank formulation:

\begin{equation}\label{lowrank}
    \Sigma = PP^T + D
\end{equation}

where $P \in \mathbb{R}^{B \times (N_w \times C) \times R}$ is the low-rank covariance factor, $R$ is the rank, and $D \in \mathbb{R}^{B \times N_w \times C}$ is the covariance diagonal.
Each parameter of the distribution, including the mean, covariance diagonal, and covariance factor, is predicted by an individual fully connected layer.
The resulting distribution is cached and used to sample multiple latent representations during training. These samples serve as stochastic predictions, contributing to the loss calculation. As training progresses, latent features are updated iteratively, and their associated distributions are refreshed accordingly, as described in Algorithm~\ref{algo:psedocodealgo}.

\subsection{Loss function}
\label{sec:lossfunction}

\subsubsection{Stochastic loss} 

To enhance the training process with uncertainty modeling, we define a stochastic loss over the classifier outputs for synthetic images. In total, we generate $\text{ipc} \times C$ synthetic images using both latent features $w_s$ and intermediate features $f_n$ from the pretrained StyleGAN-XL generator.

After computing the loss on a batch of real images, we calculate the corresponding loss $\mathcal{L}_{\text{syn}}$ for the synthetic images of the same class using a probabilistic sampling approach. Specifically, the synthetic images are passed through the same classifier as the real ones to produce output probabilities $f(I_{\text{syn}})$.
To incorporate stochasticity, we approximate the negative log-likelihood using Monte Carlo integration:

\begin{align}
 LL & = -\log p(y \mid x) \nonumber \\ 
 & = -\log \int p(y \mid f(I_{\text{syn}})) p(f(I_{\text{syn}}) \mid x) df(I_{\text{syn}}) \nonumber \\
    & \approx -\log \frac{1}{M} \sum_{m=1}^M p(y \mid f(I_{\text{syn}})_m)
\end{align}

For simplicity, we define each sample $f(I_{\text{syn}})_m$ as $q_m$, where $q_m \mid w_s \sim \mathcal{N}(\mu(w_s), \Sigma(w_s))$, based on the latent feature distribution described in Section~\ref{sec:stochastic}. We sample $M$ such instances, $\mathcal{Q} = \{ q_m \in \mathbb{R}^{\text{ipc} \times N_w \times C} \}_{m=1}^{M}$.
Since the classifier outputs $f(I_{\text{syn}}) \in \mathbb{R}^{\text{ipc} \times C}$ also carry meaningful semantic information, we combine them with the sampled predictions to construct refined probability estimates.
The refined probabilities are computed by shifting each sample by the classifier’s output: 

\begin{equation}\label{samplemc}
    q_m^* = q_m - \mu(w_s) +  f(I_{syn})
\end{equation}

\begin{figure}[t]
   \centering
      \includegraphics[width=1\linewidth]{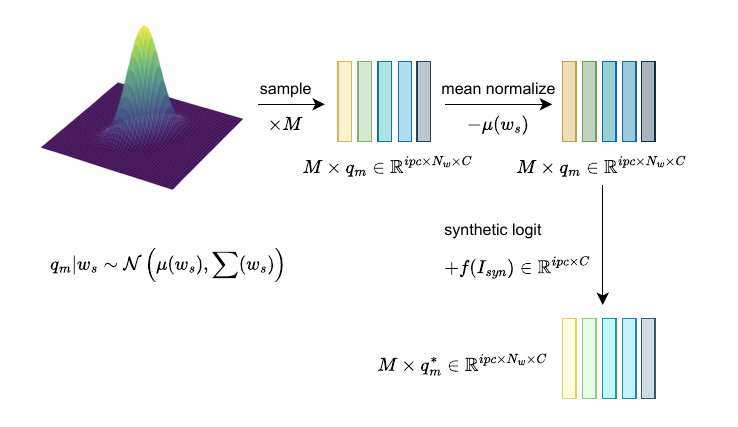}
   \caption{Illustration Eq.~\ref{samplemc} about how we sample probabilities.}
   \label{fig:stochasticdetails}
\end{figure}

This step adjusts each sample to reflect both the stochastic variability from the latent space and the semantic grounding from the classifier as in Fig.~\ref{fig:stochasticdetails}. The intuition is that the sampled probabilities encode spatial structure, while the classifier output mitigates uncertainty introduced during generation, especially at high resolutions.
We then compute a cross-entropy-based loss between these adjusted samples and the true class labels:

\begin{equation}\label{losssyn}
    \mathcal{L}_{\text{syn}}^c = -\frac{1}{L} \sum_{w=1}^{N_w} \sum_{m=1}^{M} \left( \log \sum_{m=1}^{M} \exp(y_c \cdot (q_m^*)_c) + \log M \right)
\end{equation}

Here, the label $y_c$ is repeated $M$ times to match the number of samples, $L=N_w \times M$. As in prior work~\cite{monteiro2020stochastic}, we apply the log-sum-exp trick for numerical stability.


\subsubsection{Gradient matching loss}  
To further align synthetic data with the underlying distribution of real data, we apply gradient matching during distillation. Each class is processed independently. For real images, we compute the standard cross-entropy loss, denoted $\mathcal{L}_{\text{real}}^c$, based on the classifier’s predictions and the ground truth labels.
We then explicitly compute the gradients of $\mathcal{L}_{\text{real}}^c$ and the stochastic loss $\mathcal{L}_{\text{syn}}^c$ with respect to the classifier parameters. 
This new gradient matching loss ensures that the synthetic data not only reflect class semantics but also induce training dynamics consistent with those of the original dataset.


%% file: sec/3_experiment.tex
\begin{table*}[tb]\setlength{\tabcolsep}{2pt}
   \centering
   \resizebox{1.0\textwidth}{!}{%
      \begin{tabular}{llcccccccccc}
        \toprule
          & methods & ImNet-A & ImNet-B & ImNet-C & ImNet-D & ImNet-E & ImNette & ImWoof & ImNet-Birds & ImNet-Fruits & ImNet-Cats \\
         \midrule
         \multirow{2}{*}{MTT~\cite{cazenavette2022dataset}} & pixel & 33.4$_{\pm{1.5}}$ & 34.0$_{\pm{3.4}}$ & 31.4$_{\pm{3.4}}$ & 27.7$_{\pm{2.7}}$ & 24.9$_{\pm{1.8}}$ & 24.1$_{\pm{1.8}}$ & 16.0$_{\pm{1.2}}$ & 25.5$_{\pm{3.0}}$ & 18.3$_{\pm{2.3}}$ & 18.7$_{\pm{1.5}}$ \\
          & GLaD~\cite{cazenavette2023generalizing} & 39.9$_{\pm{1.2}}$ & 39.4$_{\pm{1.3}}$ & 34.9$_{\pm{1.1}}$ & 30.4$_{\pm{1.5}}$ & 29.0$_{\pm{1.1}}$& 30.4$_{\pm{1.5}}$& 17.1$_{\pm{1.1}}$ & 28.2$_{\pm{1.1}}$ & 21.1$_{\pm{1.2}}$ & 19.6$_{\pm{1.2}}$\\
         \midrule
         \midrule
         \multirow{5}{*}{DC~\cite{zhao2020dataset}} & pixel & 38.7$_{\pm{4.2}}$ & 38.7$_{\pm{1.0}}$ & 33.3$_{\pm{1.9}}$ & 26.4$_{\pm{1.1}}$ &  27.4$_{\pm{0.9}}$ & 28.2$_{\pm{1.4}}$ & 17.4$_{\pm{1.2}}$ & 28.5$_{\pm{1.4}}$ & 20.4$_{\pm{1.5}}$ & 19.8$_{\pm{0.9}}$\\
          & GLaD~\cite{cazenavette2023generalizing} & 41.8$_{\pm{1.7}}$ & 42.1$_{\pm{1.2}}$ & 35.8$_{\pm{1.4}}$ & 28.0$_{\pm{0.8}}$ & 29.3$_{\pm{1.3}}$ & 31.0$_{\pm{1.6}}$ & 17.8$_{\pm{1.1}}$ & 29.1$_{\pm{1.0}}$ & 22.3$_{\pm{1.6}}$ & 21.2$_{\pm{1.4}}$ \\
         & SLFD & \textbf{43.27}$_{\pm{1.6}}$ & \textbf{43.31}$_{\pm{1.1}}$ & \textbf{36.76}$_{\pm{1.1}}$ & \textbf{29.26}$_{\pm{1.5}}$ & \textbf{31.02}$_{\pm{1.4}}$ & \textbf{33.10}$_{\pm{1.2}}$ & \textbf{19.13}$_{\pm{1.1}}$ & \textbf{30.22}$_{\pm{1.2}}$ & \textbf{23.62}$_{\pm{1.3}}$ & \textbf{21.65}$_{\pm{1.3}}$ \\
         \cmidrule{2-12}
         & GLaD-Conv & 44.1$_{\pm{2.4}}$ & 49.2$_{\pm{1.1}}$ & 42.0$_{\pm{0.6}}$ & 35.6$_{\pm{0.9}}$ & \textbf{35.8$_{\pm{0.9}}$} & 35.4$_{\pm{1.2}}$ & 22.3$_{\pm{1.1}}$ & 33.8$_{\pm{0.9}}$ & 20.7$_{\pm{1.1}}$ & 22.6$_{\pm{0.8}}$\\    
          & SLFD-Conv & \textbf{46.00$_{\pm{1.4}}$} & \textbf{49.96}$_{\pm{1.0}}$ & \textbf{42.60}$_{\pm{1.0}}$ & \textbf{36.40}$_{\pm{2.3}}$ & 35.44$_{\pm{0.60}}$ & \textbf{36.72}$_{\pm{1.3}}$ & \textbf{22.68}$_{\pm{0.9}}$ & \textbf{33.88}$_{\pm{1.0}}$ & \textbf{21.96}$_{\pm{0.5}}$ & \textbf{24.12}$_{\pm{1.0}}$ \\   
         \midrule
         \midrule
         \multirow{5}{*}{DM~\cite{zhao2023DM}} & pixel & 27.2$_{\pm{1.2}}$ & 24.4$_{\pm{1.1}}$ & 23.0$_{\pm{1.4}}$ & 18.4$_{\pm{1.7}}$ & 17.7$_{\pm{0.9}}$ & 20.6$_{\pm{0.7}}$ & 14.5$_{\pm{0.9}}$ & 17.8$_{\pm{0.8}}$ & 14.5$_{\pm{1.1}}$ & 14.0$_{\pm{1.1}}$\\    
         & GLaD~\cite{cazenavette2023generalizing} & 31.6$_{\pm{1.4}}$ & 31.3$_{\pm{3.9}}$ & 26.9$_{\pm{1.2}}$ & 21.5$_{\pm{1.0}}$ & 20.4$_{\pm{0.8}}$ & 21.9$_{\pm{1.1}}$ & 15.2$_{\pm{0.9}}$ & 18.2$_{\pm{1.0}}$ & 20.4$_{\pm{1.6}}$ & 16.1$_{\pm{0.7}}$ \\
         & SLFD & \textbf{39.18}$_{\pm{1.6}}$ & \textbf{38.30}$_{\pm{1.1}}$ & \textbf{32.02}$_{\pm{2.3}}$ & \textbf{25.21}$_{\pm{2.1}}$ & \textbf{23.52}$_{\pm{1.7}}$ & \textbf{28.67}$_{\pm{1.0}}$ & \textbf{17.53}$_{\pm{1.0}}$ & \textbf{24.70}$_{\pm{2.6}}$ & \textbf{23.25}$_{\pm{1.5}}$ & \textbf{18.37}$_{\pm{0.7}}$ \\
         \cmidrule{2-12}
         & GLaD-Conv & 41.0$_{\pm{1.5}}$ & 42.9$_{\pm{1.9}}$ & 39.4$_{\pm{0.7}}$ & 33.2$_{\pm{1.4}}$ & 30.3$_{\pm{1.3}}$ & 32.2$_{\pm{1.7}}$ & 21.2$_{\pm{1.5}}$ & 27.6$_{\pm{1.9}}$ & 21.8$_{\pm{1.8}}$ & 22.3$_{\pm{1.6}}$ \\    
        & SLFD-Conv & \textbf{58.72}$_{\pm{1.5}}$ & \textbf{54.04}$_{\pm{1.1}}$ & \textbf{46.56}$_{\pm{1.7}}$ & \textbf{41.88}$_{\pm{1.1}}$ & \textbf{37.68}$_{\pm{0.7}}$ & \textbf{43.72}$_{\pm{0.9}}$ & \textbf{25.44}$_{\pm{1.6}}$ & \textbf{38.08}$_{\pm{0.9}}$ & \textbf{35.44}$_{\pm{0.7}}$ & \textbf{27.56}$_{\pm{0.8}}$ \\
        \bottomrule
      \end{tabular}
    }
    \caption{Cross architecture test accuracy on ImageNet with resolution $128 \times 128$. We distill $1$ synthetic image for each class by ConvNet and train all $5$ classifiers on these $10$ synthetic images for each subset. We report the average results of $4$ unseen classifiers, ResNet18, VGG11, ViT, and AlexNet, on a real validation set to improve generalization. The rows with "*-Conv" are the results of ConvNet.}
    \label{tab:imagenet128}
   
\end{table*}

\section{Experiments}
\label{sect:experiments}

\subsection{Datasets and metrics}

We evaluate our distillation framework on three benchmark datasets: ImageNet-1K~\cite{deng2009imagenet}, CIFAR-10~\cite{krizhevsky2009learning}, and MedMNIST~\cite{yang2023medmnist}, covering both natural and medical imaging domains.
For ImageNet-1K, we conduct experiments on two configurations: $10$ class-balanced subsets at a resolution of $128 \times 128$, and $5$ subsets at a higher resolution of $256 \times 256$. Each subset contains $10$ classes which have similar categories.
CIFAR-10 is a standard benchmark for low-resolution image classification, consisting of $60,000$ images ($6,000$ per class) across $10$ categories, each image having a resolution of $32 \times 32$.

For the medical domain, we use the PathMNIST variant of the MedMNIST collection~\cite{yang2023medmnist}, which focuses on histopathology images. This dataset contains a total of $107,180$ samples, split into $89,996$ for training, $10,004$ for validation, and $7,180$ for testing. All images are resized to $28 \times 28$ to standardize input dimensions.

To benchmark our approach, we compare against recent state-of-the-art methods such as GLaD~\cite{cazenavette2023generalizing}, focusing on the downstream classification task. We report the cross-architecture test accuracy to assess generalization and robustness of the distilled datasets.

\subsection{Implementation details}
The distillation process is performed using a ConvNet architecture. Once the synthetic dataset is generated, we evaluate its effectiveness by training $5$ different classifiers, ConvNet, ResNet18, VGG11, ViT, and AlexNet, on the distilled images. These models are then tested on the real validation sets from ImageNet in all experimental settings.
Our primary evaluation metric is cross-architecture accuracy, which reflects the average test performance of the $4$ classifiers not involved in distillation (ResNet18, VGG11, ViT, and AlexNet). This setup is intended to assess the generalization ability of the distilled data across different model architectures. For completeness, we also report results when the ConvNet is used for both distillation and downstream classification.
Each reported accuracy is averaged over $5$ independent runs, and we include the corresponding standard deviation to reflect consistency and reliability.
All experiments are conducted on a single NVIDIA A100 GPU. The total runtime is approximately 9 hours, which includes the full distillation process, training, and evaluation of all $5$ classifiers at every $100$ iterations across $5$ repetition runs. This computational setup is comparable to that used in GLaD.

\subsection{Benchmark dataset results}
\subsubsection{ImageNet}

Table~\ref{tab:imagenet128} presents a comparison between our method and existing state-of-the-art techniques across $10$ subsets of the ImageNet dataset at a resolution of $128 \times 128$. Our approach shows consistent improvements in cross-architecture performance, achieving up to a $7\%$ gain when using gradient matching (DC) and up to $36\%$ with distribution matching (DM). The ConvNet classifier, which is also used during distillation, achieves performance gains of up to $7\%$ and $63\%$, particularly notable on the ImNet-Fruits subset.
In Table~\ref{tab:imagenet256}, we report results on $5$ subsets at a higher resolution of $256 \times 256$. Across all subsets, SLFD improves classification accuracy, with the ImNet-A subset showing a significant boost of $17.7\%$. Importantly, no performance degradation is observed with the increased resolution, highlighting the robustness and scalability of our method.

\begin{table}[tb]\setlength{\tabcolsep}{2pt}
   \centering
   \resizebox{1.0\columnwidth}{!}{%
      \begin{tabular}{lccccc} 
         \toprule
          & ImNet-A & ImNet-B & ImNet-C & ImNet-D & ImNet-E  \\ 
         \midrule
         DC~\cite{zhao2020dataset} & $38.3_{\pm{4.7}}$ & $32.8_{\pm{4.1}}$ & $27.6_{\pm{3.3}}$ & $25.5_{\pm{1.2}}$ & $23.5_{\pm{2.4}}$ \\
         GLaD~\cite{cazenavette2023generalizing} & $37.4_{\pm{5.5}}$ & $41.5_{\pm{1.2}}$ & $35.7_{\pm{4.0}}$ & $27.9_{\pm{1.0}}$ & $29.3_{\pm{1.2}}$ \\
         SLFD  & \textbf{44.01$_{\pm{1.7}}$} & \textbf{43.15$_{\pm{1.6}}$} & \textbf{37.28$_{\pm{1.4}}$} & \textbf{29.78$_{\pm{1.1}}$} & \textbf{31.57$_{\pm{1.0}}$}\\
        \bottomrule
      \end{tabular}
    }
   \caption{Cross architecture test accuracy on $5$ subsets of ImageNet $(256 \times 256)$ with DC method.}
   \label{tab:imagenet256}
\end{table}

\begin{figure*}[tb]
   \centering
    \includegraphics[width=1\linewidth]{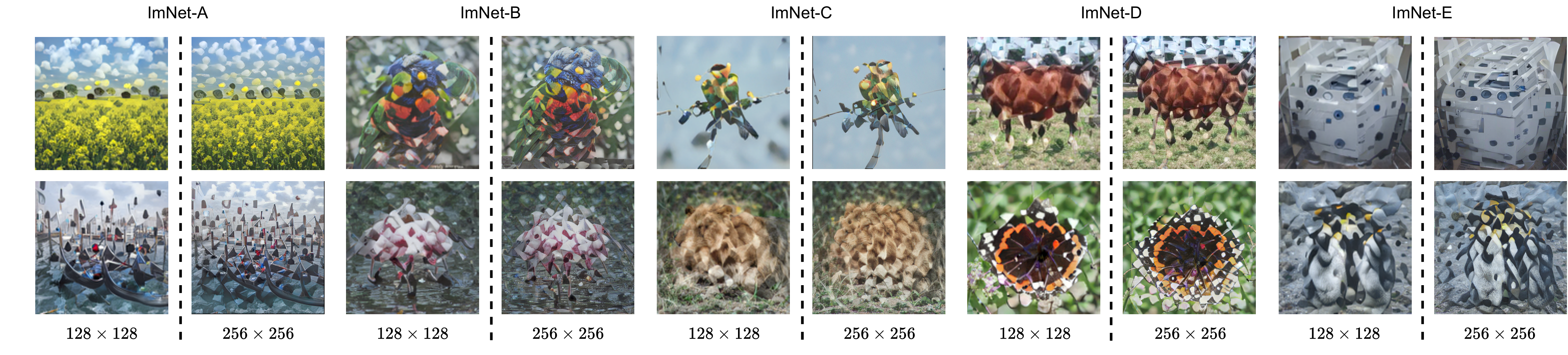}
   \caption{Example images distilled from ImageNet subsets.}
   \label{fig:qualitative_imagenet}
\end{figure*}

Figure~\ref{fig:qualitative_imagenet} presents qualitative results of synthetic images generated by our SLFD method. These samples can be broadly divided into two categories.
In the first row, the synthetic images closely reflect recognizable features from the corresponding real-world classes, capturing salient visual cues that align with human interpretation. In contrast, the second row contains examples where object-level clarity is reduced. These images often contain overlapping parts or fragmented features, for instance, multiple boats appearing merged in the \textit{Gondola} class or compressed wing patterns in the \textit{Admiral} butterfly category. Although individual objects may be harder to distinguish, higher-resolution synthesis contributes to richer visual content, revealing finer details or a greater density of object-related elements.
It is important to note that generating photorealistic images is not the primary goal of dataset distillation. Instead, the objective is to encode class-discriminative features that facilitate effective model training. While some synthetic images may appear ambiguous or unintelligible to human observers, they can still provide meaningful training signals. Often, these images prioritize texture and statistical patterns over explicit structure, especially when classes within a subset share visual similarities. This is consistent with findings in prior work~\cite{cazenavette2022dataset}, where the utility of synthetic data lies more in its representational efficiency than in its visual fidelity.



\subsubsection{CIFAR-10} 
Following our evaluation on ImageNet, we assess the effectiveness of SLFD on the CIFAR-10 dataset to test its generalization to lower-resolution data. Table~\ref{tab:cifar32} reports results across $4$ classifiers not used during distillation. We consider two generator settings: $G_r$, where the generator is randomly initialized, and $G_t$, where the generator is pretrained on the ImageNet dataset.
Across both settings, our method consistently outperforms GLaD~\cite{cazenavette2023generalizing}, demonstrating the adaptability of SLFD to different data domains and resolutions.

\begin{table}[tb]\setlength{\tabcolsep}{2pt}
    \centering
    \resizebox{1.0\columnwidth}{!}{%
      \begin{tabular}{lccccc}
        \toprule
          & average & AlexNet & ResNet18 & VGG11 & ViT  \\
         \midrule
         pixel & 26.0$_{\pm{0.4}}$ & 25.9$_{\pm{0.2}}$ & 27.3$_{\pm{0.5}}$ & 28.0$_{\pm{0.5}}$ & 22.9$_{\pm{0.3}}$ \\
         GLaD $G_r$~\cite{cazenavette2023generalizing} & 26.6$_{\pm{0.6}}$ & \textbf{30.1$_{\pm{0.5}}$} & 27.3$_{\pm{0.2}}$ & 28.0$_{\pm{0.9}}$ & 21.2$_{\pm{0.6}}$ \\
         GLaD $G_t$~\cite{cazenavette2023generalizing} & 26.3$_{\pm{0.5}}$ & 26.0$_{\pm{0.7}}$ & 27.6$_{\pm{0.6}}$ & 28.2$_{\pm{0.6}}$ & 23.4$_{\pm{0.2}}$ \\
         SLFD $G_t$ & \textbf{28.1$_{\pm{0.6}}$} & 26.8$_{\pm{0.6}}$ & \textbf{28.5$_{\pm{0.9}}$} & \textbf{28.7$_{\pm{0.4}}$} & \textbf{28.5$_{\pm{0.3}}$}\\
        \bottomrule
      \end{tabular}
    }
    \caption{Cross architecture results on CIFAR-10 with DC and random initialization $G_r$ or pre-trained $G_t$.}
    \label{tab:cifar32}
\end{table}

\subsubsection{MedMNIST}
\begin{figure*}[tb]
     \begin{subfigure}[b]{1\linewidth}
         \includegraphics[width=\linewidth]{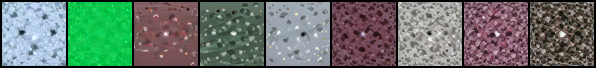}
         \caption{SLFD resolution 64}
        
     \end{subfigure}
     \hfill
     \begin{subfigure}[b]{1\linewidth}
         \includegraphics[width=\linewidth]{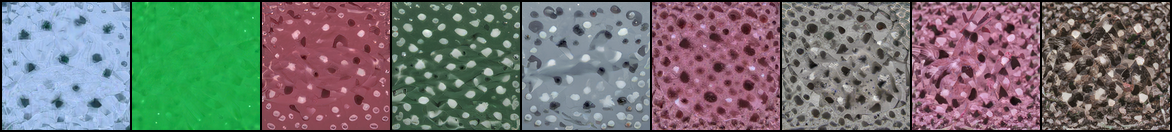}
         \caption{SLFD resolution 128}
        
     \end{subfigure}
     \begin{subfigure}[b]{1\linewidth}
         \includegraphics[width=\linewidth]{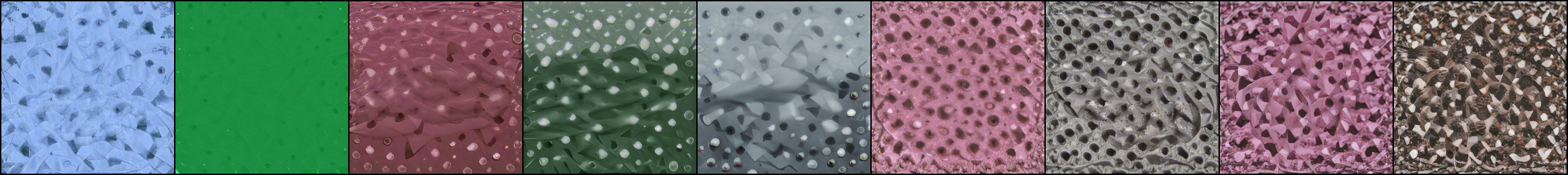}
         \caption{SLFD resolution 256}
         \label{}
     \end{subfigure}
   \caption{Synthetic images from MedMNIST~\cite{yang2023medmnist} for different output resolutions.}
\label{fig:qual_medmnist}
\end{figure*}

Building on our results from natural image datasets, we further evaluate SLFD on the MedMNIST dataset to test its effectiveness in the medical imaging domain. Table~\ref{tab:medmnist} presents the classification performance across various input resolutions. For training, we upsample the original $28 \times 28$ images to resolutions of $64$, $128$, and $256$ using bicubic interpolation.
While the performance of GLaD~\cite{cazenavette2023generalizing} slightly declines as image resolution increases from 64 to 256, our method shows a consistent improvement, demonstrating greater robustness and suitability for high-resolution medical imagery.
Figure~\ref{fig:qual_medmnist} provides qualitative examples of the generated synthetic data. The $9$ classes in the dataset are visually distinguishable, indicating that SLFD effectively captures class-specific structure. Notably, the real histology images predominantly appear in shades of pink due to hematoxylin and eosin (H\&E) staining. In contrast, our synthetic images exhibit a broader color range, including green and light blue, resulting from normalization applied during preprocessing. This variation does not hinder classification, suggesting that our method can retain discriminative features even under color shifts.

\begin{table}[tb]
   \centering
   \resizebox{0.8\columnwidth}{!}{%
      \begin{tabular}{lccc}
         \toprule
          & res=64 & res=128 & res=256  \\
         \midrule
         GLaD~\cite{cazenavette2023generalizing} & 36.88$_{\pm1.1}$ & 35.77$_{\pm1.2}$ & 35.64$_{\pm1.7}$ \\
         SLFD  & 36.67$_{\pm0.6}$ & 36.27$_{\pm1.7}$ & 37.17$_{\pm2.3}$ \\
         \bottomrule
      \end{tabular}
    }
    \caption{Cross architecture result on MedMNIST~\cite{yang2023medmnist}.}
    \label{tab:medmnist}
\end{table}

\begin{figure}[tb]
   \centering
      \includegraphics[width=1\linewidth]{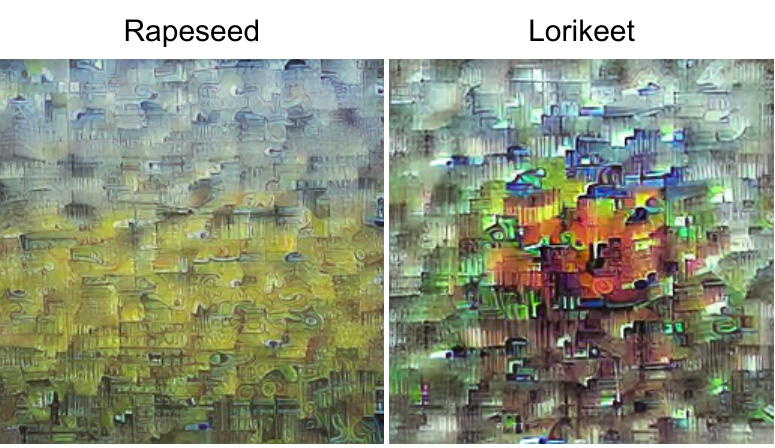}
   \caption{Results of ablation study. Distilled images with the diffusion model UViT~\cite{bao2023all} with resolution $(256\times 256)$.}
   \label{fig:ablation_diffusion}
\end{figure}

\begin{figure}[tb]
   \centering
      \includegraphics[width=1\linewidth]{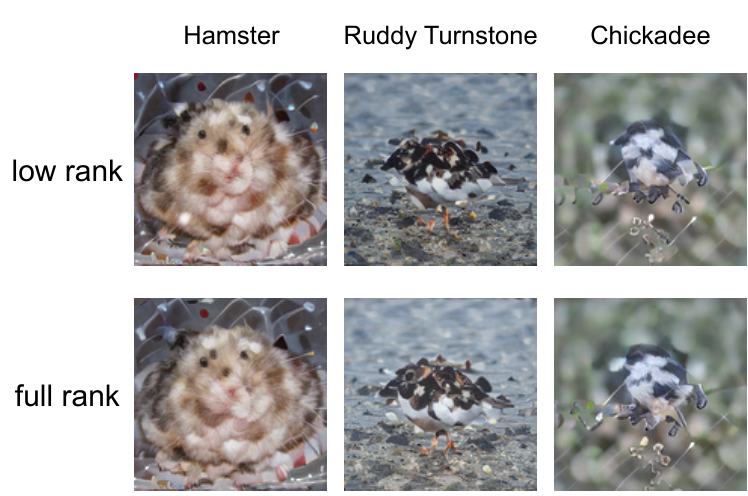}
   \caption{Results of ablation study. Qualitative examples from low-rank SLFD and full-rank SLFD on ImageNet with resolution $(128\times 128)$.}
   \label{fig:ablation_rank}
\end{figure}

\subsection{Ablation study} 

To better understand the impact of the generative backbone, we compare two generative models: StyleGAN-XL and the class-conditional diffusion model UViT~\cite{bao2023all}. The results are summarized in Table~\ref{tab:diffusion256}.
In this experiment, we integrate UViT into our distillation pipeline by replacing StyleGAN-XL with UViT as the generator. To simplify the integration, we exclude the autoencoder component of the diffusion model and adapt the remaining modules to our framework. Since UViT expects two-dimensional latent inputs, we reshape our existing 1D latent representations into a $32 \times 32$ spatial format.
After training, we use both $w_s$ and $f_n$ to generate synthetic images through UViT. All other experimental configurations are kept consistent to ensure a fair comparison.
The results show that UViT underperforms compared to StyleGAN-XL in this setting. We attribute this to the fact that the latent features generated by the mapping network are specifically structured for the StyleGAN-XL architecture, and may not transfer effectively to a diffusion-based generator.
Figure~\ref{fig:ablation_diffusion} shows qualitative examples of synthetic images generated using UViT. Notably, these images emphasize global structure, which aligns with the inductive bias of vision transformers~\cite{dosovitskiy2010image}.

\begin{table}[tb]\setlength{\tabcolsep}{2pt}
    \centering
    \resizebox{1.0\columnwidth}{!}{%
      \begin{tabular}{llccccc}
         \toprule
         & Methods & ImNet-A & ImNet-B & ImNet-C & ImNet-D & ImNet-E  \\
         \midrule
         \multirow{2}{*}{GAN} & GLaD & 37.4$_{\pm{5.5}}$ & 41.5$_{\pm{1.2}}$ & 35.7$_{\pm{4.0}}$ & 27.9$_{\pm{1.0}}$ & 29.3$_{\pm{1.2}}$ \\
         &SLFD & \textbf{44.01$_{\pm{1.7}}$} & \textbf{43.15}$_{\pm{1.6}}$ & \textbf{37.28}$_{\pm{1.4}}$ & \textbf{29.78}$_{\pm{1.1}}$ & \textbf{31.57}$_{\pm{1.0}}$\\
         \midrule
         \multirow{2}{*}{UViT} &GLaD & 37.70$_{\pm{1.5}}$ & 38.96$_{\pm{1.7}}$ & \textbf{33.11}$_{\pm{1.7}}$ & 26.38$_{\pm{1.2}}$ & 27.34$_{\pm{1.1}}$ \\
         &SLFD & \textbf{37.92}$_{\pm{2.1}}$ & \textbf{38.97}$_{\pm{1.8}}$ & 32.67$_{\pm{1.5}}$ & \textbf{26.78}$_{\pm{1.3}}$ & \textbf{28.21}$_{\pm{1.6}}$  \\
        \bottomrule
      \end{tabular}
    }  
    \caption{GAN vs. UViT~\cite{bao2023all} as backbone for $f_n$, pretrained on ImageNet.}
    \label{tab:diffusion256}
\end{table}

\begin{table}[tb]\setlength{\tabcolsep}{2pt}
   \centering
   \resizebox{1.0\columnwidth}{!}{%
      \begin{tabular}{llccccc}
        \toprule
         Res & Methods & ImNet-A & ImNet-B & ImNet-C & ImNet-D & ImNet-E  \\
         \midrule
         \multirow{3}{*}{128} & GLaD(DC)~\cite{cazenavette2023generalizing} & 41.8$_{\pm{1.7}}$ & 42.1$_{\pm{1.2}}$ & 35.8$_{\pm{1.4}}$ & 28.0$_{\pm{0.8}}$ & 29.3$_{\pm{1.3}}$  \\
         & SLFD$_{rank=10}$ & \textbf{43.27$_{\pm{1.6}}$} & \textbf{43.31$_{\pm{1.1}}$} & 36.76$_{\pm{1.1}}$ & 29.26$_{\pm{1.5}}$ & 31.02$_{\pm{1.4}}$  \\
         & SLFD$_{multi}$ & 43.21$_{\pm{1.6}}$ & 43.20$_{\pm{1.1}}$ & \textbf{36.81}$_{\pm{1.6}}$ & \textbf{29.50}$_{\pm{1.1}}$ & \textbf{31.25}$_{\pm{1.3}}$  \\
       
         \midrule
         \multirow{3}{*}{256} & GLaD(DC)~\cite{cazenavette2023generalizing} & 37.4$_{\pm{5.5}}$ & 41.5$_{\pm{1.2}}$ & 35.7$_{\pm{4.0}}$ & 27.9$_{\pm{1.0}}$ & 29.3$_{\pm{1.2}}$ \\
         & SLFD$_{rank=10}$ & 44.01$_{\pm{1.7}}$ & 43.15$_{\pm{1.6}}$ & 37.28$_{\pm{1.4}}$ & 29.78$_{\pm{1.1}}$ & 31.57$_{\pm{1.0}}$\\
         & SLFD$_{multi}$ & \textbf{44.16}$_{\pm{1.2}}$ & \textbf{43.39}$_{\pm{1.4}}$ & \textbf{37.40}$_{\pm{1.2}}$ & \textbf{30.41}$_{\pm{1.2}}$ & \textbf{32.05}$_{\pm{1.4}}$ \\
        \bottomrule
      \end{tabular}
    }
    \caption{Comparison of low-rank SLFD and full-rank SLFD using a multivariate normal distribution on ImageNet.}
    \label{tab:multivariate}
\end{table}

We further evaluate the effectiveness of our low-rank multivariate normal approximation by comparing it to a full-rank variant. As shown in Table~\ref{tab:multivariate}, our low-rank formulation achieves performance that is on par with, or in some cases exceeds, that of the full-rank counterpart across $5$ ImageNet subsets. This indicates that the low-rank approximation retains sufficient expressiveness to generate informative synthetic images while offering improved computational efficiency.
Qualitative results are presented in Fig.~\ref{fig:ablation_rank}. In some cases, such as the \textit{Hamster} class, synthetic samples appear visually similar and are difficult to distinguish, regardless of the covariance rank. The full-rank version occasionally captures finer details, for example, the eye of the \textit{Ruddy Turnstone} is more pronounced. However, the low-rank approximation also shows advantages in certain instances, such as more defined mouth features in the \textit{Chickadee} class.
These results demonstrate that the low-rank strategy maintains a strong balance between efficiency and image quality. Notably, this observation holds even at higher resolutions, such as $256 \times 256$, where detail preservation is typically more challenging.

\section{Conclusion}

In this work, we propose Stochastic Latent Feature Distillation (SLFD), a novel framework for dataset distillation that explicitly models spatial correlations through a low-rank multivariate distribution over latent features. This stochastic formulation enables the generation of informative and diverse synthetic data while preserving spatial structure.
Our experimental results show that SLFD consistently improves performance across a range of datasets and architectures, particularly in high-resolution settings where maintaining detail is crucial. Ablation studies confirm the flexibility and effectiveness of our approach under different generator types and covariance rank configurations. Furthermore, SLFD demonstrates strong performance on medical imaging tasks, providing both quantitative gains and visually interpretable results on histopathology data.
These findings highlight the potential of SLFD as a general-purpose distillation method capable of scaling across domains and resolutions while retaining efficiency and robustness.

\section*{Acknowledgments}
This work was supported by the State of Bavaria, the High-Tech Agenda (HTA) Bavaria and HPC resources provided by the Erlangen National High Performance Computing Center (NHR@FAU) of the Friedrich-Alexander-Universität Erlangen-Nürnberg (FAU) under the NHR project b180dc. NHR@FAU hardware is partially funded by the German Research Foundation (DFG) - 440719683. Support was also received from the ERC - project MIA-NORMAL 101083647. S. Cechnicka is supported by the UKRI Centre for Doctoral Training AI4Health (EP / S023283/1)

%% file: sec/X_suppl.tex
\clearpage
\setcounter{page}{1}
\maketitlesupplementary

\section{Rationale}
\label{sec:rationale}
Having the supplementary compiled together with the main paper means that:
\begin{itemize}
\item The supplementary can back-reference sections of the main paper, for example, we can refer to \cref{sec:intro};
\item The main paper can forward reference sub-sections within the supplementary explicitly (e.g. referring to a particular experiment); 
\item When submitted to arXiv, the supplementary will already included at the end of the paper.
\end{itemize}
To split the supplementary pages from the main paper, you can use \href{https://support.apple.com/en-ca/guide/preview/prvw11793/mac#:~:text=Delete%20a%20page%20from%20a,or%20choose%20Edit%20%3E%20Delete).}{Preview (on macOS)}, \href{https://www.adobe.com/acrobat/how-to/delete-pages-from-pdf.html#:~:text=Choose%20%E2%80%9CTools%E2%80%9D%20%3E%20%E2%80%9COrganize,or%20pages%20from%20the%20file.}{Adobe Acrobat} (on all OSs), as well as \href{https://superuser.com/questions/517986/is-it-possible-to-delete-some-pages-of-a-pdf-document}{command line tools}.